\documentclass[10pt,twocolumn,letterpaper]{article}

\usepackage{iccv}
\usepackage{times}
\usepackage{epsfig}
\usepackage{graphicx}
\usepackage{amsmath}
\usepackage{amssymb}
\usepackage{algorithm}
\usepackage{algorithmicx}
\usepackage{algpseudocode}
\usepackage{pifont}
\usepackage{comment}

\usepackage{caption}
\usepackage{booktabs}  
\usepackage{threeparttable}  
\usepackage{multicol}  
\usepackage{multirow}  
\usepackage{tabu} 
\usepackage{bm}
\usepackage{float}
\usepackage[normalem]{ulem}
\usepackage[british,UKenglish,USenglish,english,american]{babel}
\usepackage{tabulary}
\usepackage[dvipsnames]{xcolor}
\usepackage{soul}
\usepackage{xspace}\xspace
\usepackage{adjustbox}
\usepackage{array}
\usepackage{dblfloatfix}
\usepackage{transparent}

\usepackage{algorithm}
\usepackage{listings}
\usepackage{bm}
\usepackage{bbm}

\usepackage{caption}
\usepackage{booktabs}  
\usepackage{threeparttable}  
\usepackage{multicol}  
\usepackage{multirow}  
\usepackage{tabu} 
\usepackage{bm}
\usepackage{float}
\usepackage[normalem]{ulem}
\usepackage[british,UKenglish,USenglish,english,american]{babel}

\newlength\savewidth\newcommand\shline{\noalign{\global\savewidth\arrayrulewidth
  \global\arrayrulewidth 1pt}\hline\noalign{\global\arrayrulewidth\savewidth}}

\newcommand{\Tref}[1]{Table~\ref{#1}}
\newcommand{\Eref}[1]{Equation~(\ref{#1})}
\newcommand{\Fref}[1]{Figure~\ref{#1}}

\newcommand{\one}[1]{\mathbbm{1}_{[#1]}}

\newcommand{\cmark}{\ding{51}}%

\newcolumntype{x}[1]{>{\centering\arraybackslash}p{#1pt}}
\newcolumntype{y}[1]{>{\raggedright\arraybackslash}p{#1pt}}
\newcolumntype{z}[1]{>{\raggedleft\arraybackslash}p{#1pt}}

\definecolor{Highlight}{HTML}{39b54a}  % green

\renewcommand{\hl}[1]{\textcolor{Highlight}{#1}}

\newcommand{\hlb}[1]{\textcolor{blue}{#1}}

\newcommand{\reshl}[3]{{#1}\fontsize{6.5pt}{0.25em}\selectfont{~\hl{(${#2}$\textbf{#3})}}}

\newcommand{\reshlb}[3]{{#1}\fontsize{6.5pt}{0.25em}\selectfont{~\hlb{(${#2}$\textbf{#3})}}}
\usepackage{bm}
\usepackage{float}
\usepackage[normalem]{ulem}

\usepackage[pagebackref=true,breaklinks=true,letterpaper=true,colorlinks,bookmarks=false]{hyperref}

\iccvfinalcopy % *** Uncomment this line for the final submission

\def\iccvPaperID{1693} % *** Enter the ICCV Paper ID here

\ificcvfinal\fi

\begin{document}

%%%%%%%%% TITLE
\title{Unsupervised Finetuning}

% \and

\author{Suichan Li$^{1,}$\thanks{Equal contribution, $\dagger$ Dongdong Chen is the corresponding author}, Dongdong Chen$^{2,*,\dagger}$, Yinpeng Chen$^{2}$, Lu Yuan$^{2}$, Lei Zhang$^{2}$, Qi Chu$^{1}$, Bin Liu$^{1}$, Nenghai Yu$^{1}$, \\
$^{1}$University of Science and Technology of China 
\quad\quad $^{2}$Microsoft Cloud \& AI\\
\{lsc1230@mail., qchu@, flowice@, ynh@\}ustc.edu.cn, cddlyf@gmail.com, \\
\{ yiche, luyuan, leizhang\}@microsoft.com}

\maketitle
\ificcvfinal\thispagestyle{empty}\fi

%%%%%%%%% ABSTRACT
\begin{abstract}
This paper studies ``unsupervised finetuning'', the symmetrical problem of the well-known ``supervised finetuning''. Given a pretrained model and small-scale unlabeled target data, unsupervised finetuning is to adapt the representation pretrained from the source domain to the target domain so that better transfer performance can be obtained. This problem is more challenging than the supervised counterpart, as the low data density in the small-scale target data is not friendly for unsupervised learning, leading to the damage of the pretrained representation and poor representation in the target domain. In this paper, we find the source data is crucial when shifting the finetuning paradigm from supervise to unsupervise, and propose two simple and effective strategies to combine source and target data into unsupervised finetuning: \textbf{``sparse source data replaying''}, and \textbf{``data mixing''}.  The motivation of the former strategy is to add a small portion of source data back to occupy their pretrained representation space and help push the target data to reside in a smaller compact space; and the motivation of the latter strategy is to increase the data density and help learn more compact representation. To demonstrate the effectiveness of our proposed  ``unsupervised finetuning'' strategy, we conduct extensive experiments on multiple different target datasets, which show better transfer performance than the naive strategy.

\end{abstract}

%%%%%%%%% BODY TEXT
\section{Introduction}
In recent years, visual recognition has achieved tremendous success \cite{szegedy2015going,he2016deep,tan2019efficientnet} from the development of deep neural networks and large-scale labeled data. However, in some real applications, acquiring such large-scale data is so difficult or even possible, let alone the intensive annotation. Therefore, \emph{finetuning from pretraining} \cite{kolesnikov2019big,caron2019unsupervised,fu2020unsupervised} has become the dominant training paradigm for such application scenarios, \ie, first pretrain the model on the large-scale source data, then finetune the pretrained model with few labeled target data, which we call ``supervised finetuning''.

In this paper, we study \emph{``unsupervised finetuning''}, which shares the similar goal as supervised finetuning that tries to adapt the representation pretrained from the source domain to the target domain,  so that better downstream task performance can be achieved, \eg, retrieval, unsupervised clustering, and few-shot transfer via supervised finetuning with few labeled samples. To the best of our knowledge, ``unsupervised finetuning'' is not well studied in the computer vision field and there is almost no prior work that studies generic unsupervised finetuning, even though unsupervised pretraining is actively studied as an extremely hot topic in recent two years.

At the first glance, implementing unsupervised finetuning is very intuitive, \ie, continue unsupervised learning on the unlabeled target data from the pretrained representation, in a similar way as supervised finetuning. However, this naive way does not work that well, because existing unsupervised learning schemes (\eg., contrastive learning) often require large-scale data to work properly \cite{phoo2020self}. Otherwise, they will destroy the original pretrained representation structure and struggle in learning a compact representation in the target domain. In the following section, we will analyze the mostly used unsupervised contrastive loss, and hypothesize that this may be pre-determined by its own formulation. Inspired by the analysis, we discover two simple and effective strategies: \emph{sparse source data replaying} and \emph{data mixing}.

Specifically, for data replaying, we add some source data back, and involve both these source data and the unlabeled target data into the contrastive learning process. The main motivation here is using such source data to occupy their pretrained representation space and help push the target data to reside in a smaller compact space. However, to ensure the efficiency of unsupervised finetuning, only a small portion (e.g., 10\%) of source data is involved. To further increase the data density and help contrastive learning on a small data scale, data mixing is used to create extra positive samples. In details, given two random images, we create a mixed image and regard both the original two images as positive samples weighted by the mixing weights. It can help pull similar images closer and learn more compact representation. 

To demonstrate the effectiveness of our proposed ``unsupervised finetuning'' strategy, we conduct extensive experiments on multiple small-scale target datasets  with both unsupervised pretrained models and supervised pretrained models. We also evaluate the unsupervised finetuned model on different transfer tasks to evaluate its transfer performance, including unsupervised image retrieval, unsupervised clustering, and few-shot transfer. Experimental results show that the proposed unsupervised finetuning strategy can achieve much better performance than the naive unsupervised finetuning strategy and the original pretrained representation. 

To summarize, our contributions are in three aspects:
\begin{itemize}
    \item To the best of our knowledge, we are the first that systematically analyzed the  ``unsupervised finetuning'' problem by dissecting the widely used contrastive loss.
    \item Based on the analysis, we discover two simple and effective strategies, \ie, \emph{sparse source data replaying} and \emph{data mixing}, based on which better unsupervised finetuning can be achieved.
    \item Extensive experiments are conducted and demonstrate the superiority and generalization ability of our unsupervised finetuning strategy across different target datasets, pretraining methods and transfer tasks.
\end{itemize}

\section{Related Works}
\paragraph{Supervised pretraining.} Benefiting from the powerful learning capacity, deep neural networks revolutionized many computer vision tasks in recent years, such as image recognition \cite{he2016deep,szegedy2015going}, object detection \cite{ren2015faster,girshick2015fast,carion2020end}, and semantic segmentation \cite{chen2017deeplab,long2015fully}. However, such deep networks often require a large-scale labeled dataset for training, \eg, ImageNet\cite{deng2009imagenet} and JFT-300M \cite{sun2017revisiting}. In order to achieve great performance on small datasets, model pretraining \cite{zoph2020rethinking,zhou2020unified} has demonstrated to be an effective way, and most previous success comes from supervised pretraining. For example, the ImageNet pretrained models are widely used in a broad range of downstream tasks \cite{ren2015faster,wang2018learning,fu2020improving} and can significantly boost the downstream performance. In the recent work BiT \cite{kolesnikov2019big}, Alexander \textit{et al.} demonstrate pretraining big models on big supervised data is extremely powerful and can achieve excellent transfer performance on downstream few-shot tasks.

\paragraph{Unsupervised Pretraining.} Despite the great success of supervised pretraining, it is relatively difficult to scale up because annotating a large scale labeled data is extremely costly and time-consuming. Over the past few years, unsupervised or self-supervised pretraining has attracted more and more attentions. Early unsupervised pretraining methods often design some pretext tasks as the learning signal. Examples include recovering the input under corruptions, \eg, denoise auto-encoders\cite{vincent2008extracting}, cross-channel auto-encoders \cite{zhang2017split}. Other popular pre-tasks include rotation prediction \cite{gidaris2018unsupervised} and patch ordering \cite{doersch2015unsupervised}. Recently, contrastive learning \cite{wu2018unsupervised,he2020momentum,chen2020simple} has almost dominated this field. The underlying motivation is the InfoMax principle \cite{oord2018representation}, which aims to maximize the mutual information across different augmentations of the same images. For detailed implementation, it encourages the representation of different augmentations of the same image to be similar and that of different images to be dissimilar. Many recent unsupervised pretraining methods \cite{he2020momentum,grill2020bootstrap,caron2020unsupervised} have shown unsupervised pretraining on  ImageNet can achieve comparable transfer performance on downstream datasets to the supervised pretraining counterpart. But since unsupervised learning starts \emph{tabula rasa}, it still requires extremely large amounts of unlabeled data to learn a good representation, which is also illustrated in the work \cite{li2020fewer,phoo2020self}.

Nevertheless, in many application scenarios, it is difficult or even impossible to collect a large-scale dataset, such as finegrained classification and some other medical related tasks, especially when the semantic category number is small. Inspired by it, we study ``unsupervised finetuning'', which aims to relax this requirement and studies how to finetune the pretrained model with the small-scale target data in an unsupervised way. Our method is orthogonal to both supervised and unsupervised pretraining works. On the one hand, our unsupervised finetuning can work upon both supervised and unsupervised pretrained models. On the other hand, our unsupervised finetuning is essentially contrastive learning but with two new strategies, so integrating better contrastive learning methods can also improve our unsupervised finetuning method.

\paragraph{Target Domain Transfer.} Given the pretrained model, the most widely used transfer strategy is supervised finetuning, i.e., collect some labeled samples in the target domain and finetune the pretrained model in a supervised way with a proper (often smaller) learning rate.  In this paper, we study how to finetune the pretrained model with small-scale unlabeled target data, which is useful for many real application scenarios. One typical example is that, in some large-scale retrieval systems, there is only unlabeled target data. Another typical example is to provide better representation to conduct unsupervised clustering. But due to the lack of label supervision, unsupervised finetuning is not a trivial task like supervised finetuning. Using a small number of unlabeled target data, naively unsupervised learning from the pretrained model is not satisfactory. Because it will easily destroy the original representation structure and degrade to a trivial solution. In the experiment part, to demonstrate the better transfer performance from unsupervised finetuned representation, we will use several typical transfer strategies in the target domain for evaluation, including few-shot transfer,  unsupervised retrieval and clustering. 
\section{Motivation}

Given a small scale of unlabeled target data $T=\{x^t_1,...,x^t_n\}$, there are two naive ways to learn representation from these data. One is directly conducting unsupervised learning on this data from scratch, and another is to conduct unsupervised learning on this data from the pretrained representation that is trained on other source domain. To implement the first solution, we adopt the existing state-of-the-art unsupervised learning method MoCoV2\cite{chen2020improved} on two example target datasets: Caltech101~\cite{fei2004learning} and Pet~\cite{parkhi2012cats}, which contain  a total of 3680, 3060 images for training respectively. To evaluate the quality of the learned representation, we finetune the pretrained model with a few labeled samples to get the few-shot transfer performance. As the reference, we also show the few-shot transfer performance of the unsupervised pretrained model from the large-scale source dataset, \ie, ImageNet. 

As shown in the second and third row of \Tref{tbl:motivation}, we can observe that conducting unsupervised learning directly on the small-scale target dataset cannot learn a good representation, and its few-shot transfer performance is much worse than the unsupervised pretrained model on the large-scale source dataset (ImageNet). But from another perspective, it demonstrates the value of the second baseline, \ie, finetuning upon the pretrained representation from the source domain. However, how to conduct unsupervised finetuning is still unknown. As the simple baseline, we follow a similar strategy as the well studied ``supervised finetuning''. In other words, we continue unsupervised learning on the target data with a smaller learning rate by using the pretrained model as the initialization. The corresponding transfer performance is shown in the last row of \Tref{tbl:motivation}. It can be seen that, on Caltech101, the transfer performance is slightly improved, but on Pet37, the transfer performance becomes even worse. It means unsupervised finetuning is not as trivial as simple finetuning. 

To further understand why unsupervised finetuning is nontrivial, we follow the analysis in the work \cite{chen2020intriguing} about the contrastive loss, which represents the generalized contrastive loss in the below form:
\begin{equation}
\label{eq:gc}
\mathcal{L}_{\text{contr}} = \mathcal{L}_{\text{align}} + \lambda \mathcal{L}_{\text{distr}}
\end{equation}
where $\mathcal{L}_{\text{align}}$ is the alignment term that encourages the learned representation of different augmented views of the same image to be as close as possible. And $\mathcal{L}_{\text{distr}}$ is the distribution term that encourages the learned representation to match a prior distribution of high entropy. Following the definition in \cite{chen2020simple}, the standard contrastive loss is  defined as:
\begin{equation}
\label{eqn:contrastive}
    \mathcal{L}_{contr} = -\frac{1}{N}\sum_{i,j\in \mathcal{MB}}\log \frac{\exp(\bm z_i\cdot\bm z_j/\tau)}{\sum_{k=1}^{2N} \one{k \neq i}\exp(\bm z_i\cdot\bm z_k/\tau)}
\end{equation}
where $\bm z_i, \bm z_j$ are the representations of the two augmented views of the same example. $\mathrm{sim}(\bm u, \bm v)$ is the cosine similarity between $\bm u$ and $\bm v$, and $\tau$ is the temperature hyper-parameter. $\mathcal{MB}$ is the mini-batch that consists of random augmented pairs of images and $N$ is the mini-batch size. In the detailed implementation, all the learned representation $\bm z_i$ will be normalized before passing into the contrastive loss. The above equation can be further rewritten into:
\begin{equation}
\begin{aligned}
\label{eqn:split_contastive}
 \mathcal{L}_{contr} &= -\frac{1}{N\tau}\sum_{i,j}\mathrm{sim}(\bm z_i, \bm z_j)
    + \\&\frac{1}{N}\sum_i\log\sum_{k=1}^{2N} \one{k \neq i}\exp(\mathrm{sim}(\bm z_i, \bm z_k)/\tau)
\end{aligned}
\end{equation}
The first and second term correspond to the alignment term $\mathcal{L}_{\text{align}}$ and distribution term in \Eref{eq:gc} respectively. More interestingly, the second term is related to pairwise potential in a Gaussian kernel and can be minimized with a uniform encoder. It will encourage the learned representation to be uniformly distributed on the hypersphere. Considering $\bm z_i$ often has a high dimension for larger representation capacity, if the training data has a limited scale, the data density is not enough to learn a compact representation because of the $\mathcal{L}_{distr}$ term. In this case, contrastive learning indeed degrades to a trivial solution, \ie, the learned target representation scatters in the whole space sparsely. This explains why  the first naive way does not work very well. 

\begin{table}[!t]
%\small
\setlength{\tabcolsep}{0.85mm}{
\begin{tabular}{c|c c c|c c c}
\hline
 \multirow{2}{*}{Model} &  \multicolumn{3}{c|}{Pet37} & \multicolumn{3}{c}{Caltech101}  \\ \cline{2-7} & 2-Shot & 4-Shot & 8-Shot & 2-Shot & 4-Shot & 8-Shot \\
 \hline
 Unsup-$T$ &  9.08 & 10.97 & 12.64 &  18.69 & 27.17 &38.27  \\
 \hline
Unsup-$S$ &   56.84 & 68.94 & 75.10 & 63.12 & 76.54 & 84.45 \\
 \hline
Unsupft-$T$ &  55.94 & 67.03 & 72.92 & 68.48 & 78.18 & 82.03 \\
 \hline
\end{tabular}
\caption{Few-shot transfer performance on two small-scale target datasets with different baseline models. Unsup-$T$ and Unsup-$S$ mean the unsupervised pretrained model on the small target dataset $T$ and the large source dataset ImageNet $S$ from scratch respectively. Unsup-$T$ means the naive unsupervised finetuned model on $T$ from the pretrained Unsup-$S$.}
\label{tbl:motivation}
}
\end{table}

\begin{figure*}[t]
\begin{center}
\includegraphics[width=1.0\linewidth]{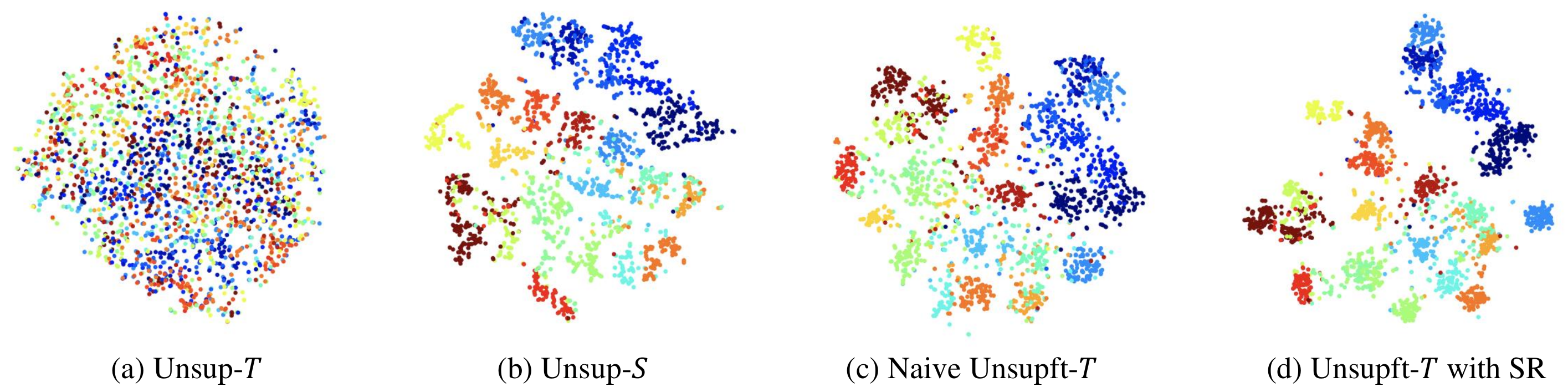}
\caption{The feature distribution visualization of the Pet dataset with different models via TSNE. (a) is the unsupervised pretrained model only on the small-scale Pet dataset from scratch, (b) is the unsupervised pretrained model on the large-scale ImageNet dataset, (c) is the naive unsupervised finetuned model on the Pet dataset by initializing with model (b), and (d) is the unsupervised finetuned model with our source data replaying.}
\label{fig:vis_cmp}
\end{center}
\end{figure*}

Conversely, if the training data has a very large scale and data density, different samples will squeeze each other, thus pushing the similar samples to be closer and learned representation more compact. Therefore, the ImageNet unsupervised pretrained model has a relatively good transfer performance. As for the naive simple unsupervised finetuning baseline, since the source data is not involved in the finetuning stage, there is much freedom for the sparse target data to wander in the whole space. This will not only easily destroy the original space structure but also struggle in learning a more compact representation. 

Trying to verify the above analysis, we further visualize the feature distribution on the Pet dataset with different pretrained models via TSNE in \Fref{fig:vis_cmp}. The visualization results roughly align with our above analysis and the few-shot transfer performance shown in the \Tref{tbl:motivation}. In details, unsupervised learning directly on the small-scale target data almost learns a uniformly distributed representation, and the naive unsupervised finetuning strategy does not improve the pretrained representation a lot either. Accompanying the visualization, we also evaluate the feature clustering quality on the source ImageNet dataset and target Pet dataset. Before and after the unsupervised finetuning, the clustering accuracy on ImageNet and Pet are $36.85\%, 47.72\%$ and $16.60\%, 47.31\%$ respectively. The significant clustering accuracy drop on the ImageNet also indicates the naive unsupervised finetuning will destroy the pretrained representation structure on the source dataset.

\section{Method}
Based on the above analysis, we discover two simple and effective strategies: \textbf{\textit{sparse source data replaying}} and \textbf{\textit{data mixing}}. We find both of them can significantly improve unsupervised finetuning for better representation learning by adding more push force from the $\mathcal{L}_{distr}$ term and more pull force from the $\mathcal{L}_{\text{align}}$ term respectively. In this paper, we implement unsupervised finetuning on top of the state-of-the-art unsupervised learning framework MoCOV2 \cite{chen2020improved}. Before introducing the above two strategies, we will first briefly introduce MoCoV2 for better understanding. 

In MoCoV2, for each image within the mini-batch, two different augmentations will be conducted. One augmented sample is regarded as query $\bm{q}$ and the other is regarded as the positive key $\bm{k}^+$. All augmented versions of other images are regarded as negative keys $\{\bm{k}_i^-\}$. Then for each image, the contrastive loss is defined as:
\begin{equation}
\label{eqn:contrastive}
    \mathcal{L}_{\bm{q}} = - \text{log}\frac{\text{exp}(\bm{\bm{z}_{\bm{q}}}\cdot\bm{z}_{\bm{k}}^+ / \tau)}{\text{exp}(\bm{\bm{z}_{\bm{q}}}\cdot\bm{z}_{\bm{k}}^+/\tau) + \sum_{i=0}^{K-1} \text{exp}(\bm{\bm{z}_{\bm{q}}}\cdot\bm{z}_{\bm{k_i}}^-/\tau)},
\end{equation}
where $\bm{z}_{\bm{q}},\bm{z}_{\bm{k}}$ are the $L_2$-normalized feature embedding by feeding $\bm{q},\bm{k}$ into two separated encoders: a query encoder and a key encoder. The query encoder and key encoder share the same architecture, \ie, the backbone of the target network followed by an extra MLP-based projection head, but have different weights. The key idea of MoCo is making the weights of the key encoder be the momentum moving average of query encoder.

\paragraph{Sparse Source Data Replaying.} To alleviate the sparse target data density issue and avoid destroying the original pretrained representation, we propose to add some source data back and mix them with the target data together for the unsupervised finetuning. The main motivation is from two aspects:
\begin{itemize}
    \item These source data can help occupy their original feature space, which leaves a smaller remaining space for the target data to reside and avoids their original representation being destroyed.
    \item The training data density is increased, thus more push force from the source data will be imposed onto the target data via the distribution term $L_{distr}$. 
    %Besides, with strong augmentations used in the contrastive learning, the space of the similar source samples will overlap with the space of the corresponding target samples and create a more compact sub-space. 
\end{itemize}
However, since the source data has a very large scale, involving all of them into the unsupervised finetuning process will significantly hurt the efficiency. Fortunately, as the source data is often well clustered in the pretraining representation space, we find that only randomly sampling a small portion (10\% by default) of source data as the representative samples is good enough, especially when equipped with the following the data mixing strategy. Thus we call this strategy ``sparse source data replaying''. As shown in \Fref{fig:vis_cmp} (d), with this strategy, a more compact target representation can be learned. The corresponding clustering accuracy on the ImageNet and Pet dataset also increases to $28.14\%, 56.11\%$ respectively, indicating better preserving of the source representation and more compact target representation.

\paragraph{Contrastive learning with Data Mixing}
To further increase the data density, we propose to use data mixing to create extra positive samples, which link different images with more alignment terms. In our method, we use cutmix\cite{yun2019cutmix} as the default instantiation of data mixing, and we empirically observed that other mixing strategies like MixUP can also work well. In details, given two images, we denote one augmented version of them as the querys $q_a$, $q_b$, and another augmented version as the positive keys $k_a^+$, $k_b^+$, then we create a mixed query sample $q_m$:
\begin{equation}
    q_m \leftarrow \textbf{M} \odot q_a + (\textbf{1} - \textbf{M}) \odot q_b
\end{equation}
when $\textbf{M} \in [0,1]^{W \times H}$ is a binary mixing weight mask, \textbf{1} is a binary mask filled with ones. $\odot$ means the element-wise multiplication operation and $W,H$ is the width and height of training images. 
Then the mixed query $q_m$ is regarded as the positive sample both for $k_a^+$ and $k_b^+$, thus the contrastive loss with respect to $q_m$ is changed to:
% \small{
\begin{equation}
\small
\begin{aligned}
\label{eqn:contrastive}
    \mathcal{L}_{q_m} & = - \Big(\lambda \text{log}\frac{\text{exp}(\bm{\bm{z}_{q_{m}}}\cdot\bm{z}_{\bm{k_a}}^+ / \tau)}{\text{exp}(\bm{\bm{z}_{\bm{q_m}}}\cdot\bm{z}_{\bm{k_a}}^+/\tau) + \sum_{i=0}^{K-1} \text{exp}(\bm{\bm{z}_{\bm{q_m}}}\cdot\bm{z}_{\bm{k_i}}^-/\tau)} \\
     + (&1-\lambda) \text{log}\frac{\text{exp}(\bm{\bm{z}_{q_{m}}}\cdot\bm{z}_{\bm{k_b}}^+ / \tau)}{\text{exp}(\bm{\bm{z}_{\bm{q_m}}}\cdot\bm{z}_{\bm{k_b}}^+/\tau) + \sum_{i=0}^{K-1} \text{exp}(\bm{\bm{z}_{\bm{q_m}}}\cdot\bm{z}_{\bm{k_i}}^-/\tau)}\Big)
\end{aligned}
\end{equation}
% }
where $\lambda$ is the combination ratio between two images and defined as the area ratio of $q_a$ in $\textbf{M}$. During training, $\lambda$ is sampled from the beta distribution Beta($\alpha$, $\alpha$), and $\alpha=1$ by default. At each training iteration, we randomly selected two training samples from the mini-batch and generate the mixed query with a sampled $\lambda$.

To understand why data mixing can help, we can also rewrite the \Eref{eqn:contrastive} into the format of \Eref{eq:gc} in a similar way. Then we will get two alignment terms bridged by the common query image $q_m$. If $q_a$ and $q_b$ are from the source and target domain respectively, the alignment terms will help pull their similar images closer and help encourage the learned representation more compact.

\begin{table*}[!th]
\small
  \begin{center}
  \setlength{\tabcolsep}{3.6mm}{
    % \begin{tabular}{l|l|cccccccc}  
    \begin{tabular}{l|l|lllll}  
    \toprule  
    & Method 
   &   Caltech101 & Pet  & CIFAR100  & Food101 & Avg\cr 
     \hline 
    \multirow{3}{*}{1-Shot Acc} & Unsup-$S$ 
     & 45.49   & 41.23   & 13.65 & 10.09 & 27.62 \cr
    & + Naive Unsupft-$T$ & \reshlb{55.65}{+}{10.16} & \reshlb{40.01}{-}{1.22}  & \reshlb{20.83}{+}{7.18} & \reshlb{28.36}{+}{18.27} & \reshlb{36.21}{+}{8.59} \cr
    & + Our Unsupft-$T$ 
    & \reshl{74.33}{+}{28.84} & \reshl{60.35}{+}{19.12}  & \reshl{33.75}{+}{20.10}  & \reshl{48.89}{+}{38.80} & \reshl{54.33}{+}{26.71} \cr
    \hline
    \multirow{3}{*}{2-Shot Acc} & Unsup-$S$ 
    & 63.12  & 56.84  & 22.95 & 15.80 & 39.68 \cr
    & + Naive Unsupft-$T$ & \reshlb{68.48}{+}{5.36}  & \reshlb{55.94}{-}{0.9} & \reshlb{30.15}{+}{7.20} & \reshlb{39.64}{+}{23.84} & \reshlb{48.55}{+}{8.87}\cr
    & + Our Unsupft-$T$ 
    & \reshl{81.02}{+}{17.88}  & \reshl{70.29}{+}{13.45}  & \reshl{44.51}{+}{21.56}  & \reshl{61.99} {+}{46.19}  & \reshl{64.45} {+}{24.77} \cr 
    \hline
    \multirow{3}{*}{4-Shot Acc} & Unsup-$S$   
     & 76.54  & 68.94    & 35.08  & 25.49  & 51.51\cr
    & + Naive Unsupft-$T$ & \reshlb{78.18}{+}{1.64}  & \reshlb{67.03}{-}{1.91} & \reshlb{39.70}{+}{4.62} & \reshlb{50.82}{+}{25.33} & \reshlb{58.93}{+}{7.42}\cr
    & + Our Unsupft-$T$ 
    & \reshl{85.35}{+}{8.81} & \reshl{76.89}{+}{7.95}  & \reshl{52.60}{+}{17.52} &  \reshl{71.14}{+}{45.65} &   \reshl{71.50}{+}{19.99}\cr
    \hline
    \multirow{3}{*}{8-Shot Acc} & Unsup-$S$ 
     & 84.45  & 75.10  & 47.84  & 35.50 &  60.72 \cr
    & + Naive Unsupft-$T$ & \reshlb{82.03}{-}{2.42} & \reshlb{72.92}{-}{2.18} & \reshlb{48.59}{+}{0.75} & \reshlb{59.58}{+}{24.08} & \reshlb{65.78}{+}{5.06}\cr
    & + Our Unsupft-$T$ 
     & \reshl{87.24}{+}{2.79} & \reshl{79.88}{+}{4.78}  & \reshl{59.68}{+}{9.14}  & \reshl{76.86}{+}{41.36} &  \reshl{75.92}{+}{15.20}  \cr
     \hline\hline
    \multirow{3}{*}{RetrievalAcc} & Unsup-$S$ 
     & 83.14  & 69.70  &  53.25 & 35.91 &  60.50 \cr
    & + Naive Unsupft-$T$ & \reshlb{84.50}{+}{1.36} & \reshlb{63.73}{-}{5.97} & \reshlb{64.27}{+}{11.02} & \reshlb{61.26}{+}{25.35} & \reshlb{68.44}{+}{7.94}\cr
    & + Our Unsupft-$T$ 
     & \reshl{85.95}{+}{2.81} & \reshl{71.67}{+}{1.97}  & \reshl{68.83}{+}{15.58}  & \reshl{75.13}{+}{39.22} &  \reshl{75.40}{+}{14.90}  \cr
     \hline\hline
    \multirow{3}{*}{ClusterAcc} & Unsup-$S$ 
     & 56.16  &  47.72 & 23.91  &  11.85 &  34.91  \cr
    & + Naive Unsupft-$T$ & \reshlb{61.33}{+}{5.17} & \reshlb{47.31}{-}{0.41} & \reshlb{41.72}{+}{17.81} & \reshlb{34.18}{+}{22.33} & \reshlb{46.14}{+}{11.23}\cr
    & + Our Unsupft-$T$ 
     & \reshl{69.93}{+}{13.77} & \reshl{56.11}{+}{8.39}  & \reshl{44.92}{+}{21.01}  & \reshl{51.74}{+}{39.89} &  \reshl{55.68}{+}{20.77}  \cr
     \bottomrule  
    \end{tabular} }
    \captionof{table}{\small Transfer performance comparison on different target datasets. ``Unsup-$S$'' denote the few-shot results by using ImageNet unsupervised pretrained model respectively, ``Naive Unsupft-$T$, and ``Our Unsup-T'' denote the few-shot results by using the naive unsupervised finetuned model and our proposed unsupervised finetuned model from ``Unsup-$S$''. It shows our unsupervised finetuning strategy can learn better representation upon the pretrained models and outperforms the naive baseline strategy. } 
    \label{tab:main_results}
     \end{center}
    % \vspace{-2em}
\end{table*} 

\section{Experiments}
In this section, we conduct extensive experiments and ablations to demonstrate the effectiveness of unsupervised finetuning in improving the transfer performance. We leverage ImageNet as the large-scale source data for pretraining and multiple public small datasets as target domain for unsupervised finetuning and transfer, including Caltech101~\cite{fei2004learning}, Pet~\cite{parkhi2012cats}, CIFAR100~\cite{krizhevsky2009learning}, and Food101~\cite{bossard2014food}. For these target datasets, we use their original data splits, which have a total of 3060, 3680, 50000, 75750 images for training respectively. Among them, Food101 (fine-grained food dataset) has a relatively large domain gap with ImageNet. To reduce the influence from randomness, each experiment is repeated three trials and reports the averaged result. For comparison, we follow the standard evaluation setting and report the mean class accuracy for Pet, Caltech101 and top-1 accuracy for other datasets. Regarding the pretrained model for unsupervised finetuning, we tried both supervised pretraining and unsupervised pretraining to demonstrate the generalization ability. 

\paragraph{Evaluation Metrics.} To evaluate the representation quality, we consider three different evaluation metrics: \emph{few-shot transfer},  \emph{retrieval accuracy}, and \emph{clustering accuracy}. The first metric requires label samples while the latter two metrics are purely unsupervised metrics. More specifically, for few-shot transfer, we randomly choose $K$ labeled samples for each category and finetune the learned representation with such labeled samples in a supervised way. 
%For semi-supervised transfer, we leverage the state-of-the-art semi-supervised learning method FixMatch \cite{sohn2020fixmatch} and utilize both the few-labeled data and all the unlabled data. 
For retrieval accuracy, we regard each image in the test set as the query, and retrieve the top-5 similar images from the training set by using the learned representation. Then the ratio of images that have the same label as the query image is defined as the retrieval accuracy. For clustering accuracy, we directly run K-Means clustering on the training set and adopt the BCubed Precision \cite{amigo2009comparison} as the clustering accuracy. Due to the space limit, we adopt the few-shot transfer performance as the main evaluation metric. In the ablation study , we will further try to apply our unsupervised finetuned representation to help semi-supervised learning.

\paragraph{Implementation details.} 
For unsupervised pretraining, we use MocoV2\cite{chen2020improved} and follow its training protocol. The initial learning rate is $0.24$ with a cosine scheduler and the batch size is 2,048. All the pretraining models are trained with 800 epochs. For supervised pretraining, we directly use the pretrained models from \cite{kolesnikov2019big}. For unsupervised finetuning, the initial learning rate is $0.015$ with batch size 256. All the models are finetuned with 200 epochs with a cosine learning rate scheduler. For transfer,  we finetune the model for $30$ epochs and the learning rate for the newly added \textit{FC} layer and pretrained layers are $3.0$ and $0.0001$ respectively, both for few-shot transfer and semi-supervised transfer. The default backbone is ResNet-50~\cite{he2016deep}.

\subsection{Main Results}
In our main experiments, we use the ImageNet unsupervised pretrained model as the initialization representation, and apply the naive unsupervised finetuning strategy and our proposed unsupervised finetuning strategy on top of it, respectively. To compare the representation quality, few-shot transfer accuracy, retrieval accuracy and unsupervised clustering accuracy are all evaluated. The detailed comparison results are shown in the \Tref{tab:main_results}. It can be seen that, by unsupervised finetuning the pretrained representation on the small-scale unlabeled target data, it can significantly improve the transfer performance in the target domain. And compared to the naive unsupervised finetuning strategy, our proposed unsupervised finetuning strategy consistently outperforms the original pretrained model and brings much larger performance gain, while the naive strategy achieves lower transfer performance than the pretrained model on  the Pet dataset. By carefully studying the above results, we can draw more interesting conclusions: 
\begin{itemize}
    \item In general, the fewer the labeled samples, the bigger performance gain unsupervised finetuning will bring. This is because finetuning with fewer labeled samples is much easier to overfit, thus having a higher requirement of the initialization representation.
    \item If the target domain has more unlabeled data (\eg, Food101), unsupervised finetuning will bring better performance and even the naive unsupervised finetuning strategy can get very plausible results.
    \item When the domain gap between the source dataset and target dataset is small, the performance gain from unsupervised finetuning is relatively smaller. For example, in the ImageNet dataset, there are a lot of animal images (especially ``dog''), which potentially have a lot of overlapped categories with the Pet dataset. Therefore, on the Pet dataset, the pretrained representation already has very good transfer performance.  
\end{itemize}

\begin{table*}[!t]
\small
  \begin{center}
  \setlength{\tabcolsep}{4.1mm}{
    % \begin{tabular}{l|l|cccccccc}  
    \begin{tabular}{l|l|lllll}  
    \toprule  
   $K$ & Method 
   &   Caltech101 & Pet  & CIFAR100  & Food101 & Avg\cr 
     \hline 
     \multirow{2}{*}{1-Shot Acc} & Sup-$S$ 
    & 45.78 & 50.04  & 25.73 & 16.48 & 38.39 \cr
    % & + Naive Unsupft-$T$ & 6.81 & 3.98 & - & - & 2.91 & \cr
    & + Our Unsupft-$T$ & \reshl{73.35}{+}{27.57}  & \reshl{61.30}{+}{11.26}  &\reshl{32.29}{+}{6.56}  &  \reshl{42.60}{+}{26.12} & \reshl{56.48}{+}{18.09}\cr
    \hline
    \multirow{2}{*}{2-Shot Acc} & Sup-$S$ 
    & 61.07  & 63.95 & 37.40 & 24.99 & 51.14 \cr
    % & + Naive Unsupft-$T$ & 9.67 & 4.97 & - &- & 3.89 &  \cr
    & + Our Unsupft-$T$ & \reshl{79.12}{+}{18.05} & \reshl{73.59}{+}{9.64}   & \reshl{45.93}{+}{8.53}   & \reshl{61.70}{+}{36.71} & \reshl{68.38}{+}{17.24} \cr 
    \hline
    \multirow{2}{*}{4-Shot Acc} & Sup-$S$   
     & 73.89  & 76.08     & 47.79  & 35.26 & 62.41   \cr
    % & + Naive Unsupft-$T$ & 14.34 & 5.39 & - & & 4.87 & \cr
    & + Our Unsupft-$T$ 
     & \reshl{84.42}{+}{10.53} & \reshl{84.41}{+}{8.33}& \reshl{55.09}{+}{7.30}  & \reshl{67.33}{+}{32.07} & \reshl{75.89}{+}{13.48}   \cr
    \hline
    \multirow{2}{*}{8-Shot Acc} & Sup-$S$ 
     & 83.87  & 82.87   & 57.46  & 45.60 & 70.90  \cr
    % & + Naive Unsupft-$T$ & 20.24 & 7.19 & - & & 66.16 & \cr
    & + Our Unsupft-$T$ 
     & \reshl{88.08}{+}{4.21}  & \reshl{85.61}{+}{3.74}  & \reshl{60.54}{+}{3.08} & \reshl{69.54}{+}{23.94} & \reshl{79.12}{+}{8.22}  \cr
    \bottomrule  
    \end{tabular} }
    \captionof{table}{\small Few-shot transfer performance on different target datasets when applying unsupervised finetuning upon the supervised pretrained models, which demonstrates the generalization ability of unsupervised finetuning. Here, ``Sup-$S$'' denote the few-shot results by using the original ImageNet supervised pretrained model, ``Our Unsup-T'' denote the few-shot results by our proposed unsupervised finetuned model from ``Sup-$S$''.} 
    \label{tab:suppre-ab}
     \end{center}
    % \vspace{-2em}
\end{table*}

\begin{table*}[!t]
\small
  \begin{center}
  \setlength{\tabcolsep}{4.0mm}{
    \begin{tabular}{l|l|lllll}  
    \toprule  
   $K$ & Method 
  & Caltech101  & Pet & CIFAR100   & Food101  & Avg\cr 
    \hline 
    \multirow{2}{*}{1-Shot Acc} 
    & Sup-$S$  
    & 53.28  & 50.67 & 26.76    & 17.05    &   36.94      \cr 
    & + Our Unsupft-$T$   
      & \reshl{74.03}{+}{20.75}  & \reshl{64.02}{+}{13.35} & \reshl{34.58}{+}{7.82}    & \reshl{42.99}{+}{25.94}   & \reshl{53.91}{+}{16.97}           \cr 
    \hline
    \multirow{2}{*}{2-Shot Acc} 
     & Sup-$S$ 
    & 66.32   & 65.96  & 37.20    & 26.20     & 48.92        \cr 
    & + Our Unsupft-$T$ 
    & \reshl{83.59}{+}{17.27}   & \reshl{79.37}{+}{13.41}      & \reshl{50.20}{+}{13.00}   & \reshl{64.33}{+}{38.13}   & \reshl{69.37}{+}{20.45}         \cr 
    \hline
    \multirow{2}{*}{4-Shot Acc}  
   & Sup-$S$  
     & 78.08  & 79.50    & 49.20    & 36.38    & 60.79      \cr 
    & + Our Unsupft-$T$  
    & \reshl{87.63}{+}{9.55} & \reshl{86.15}{+}{6.65}  & \reshl{58.32}{+}{9.12}  & \reshl{70.02}{+}{33.64}     & \reshl{75.53}{+}{14.74}         \cr 
    \hline
    \multirow{2}{*}{8-Shot Acc} 
    & Sup-$S$  
     & 86.23  & 82.69 & 59.36    & 45.84    & 68.53         \cr 
    & + Our Unsupft-$T$ 
     & \reshl{89.65}{+}{3.42} & \reshl{86.32}{+}{3.63} & \reshl{63.97}{+}{4.61}  & \reshl{71.95}{+}{26.11}  & \reshl{77.97}{+}{9.44}        \cr 
   
    \bottomrule  
    \end{tabular} }
    \captionof{table}{\small Few-shot transfer performance on the large supervised pretrained model ResNet101. The annotations are the same as \Tref{tab:suppre-ab}, and the only difference is the larger pretrained  model. It demonstrates that our unsupervised finetuning is also general to large pretrained models and brings better transfer performance.} 
    \label{tab:simple_overall_results_res101}
     \end{center}
    \vspace{-2em}
\end{table*}

\subsection{Ablation Study}
In this section, we conduct a series of ablation studies to demonstrate the generalization ability of unsupervised finetuning and the importance of our discovered strategies.

\paragraph{Generalization Ability to Supervised Pretrained Models.} Besides applying to the default unsupervised pretrained models, our proposed unsupervised finetuning strategy is also general to supervised pretrained models. In this experiment, we use the supervised pretrained ResNet50 model from BiT \cite{kolesnikov2019big} as the instantiation and apply unsupervised finetuning in a similar way. The few-shot transfer performance before and after unsupervised finetuning is shown in the \Tref{tab:suppre-ab}. It can be seen that, our unsupervised finetuning can also boost the transfer performance of supervised pretrained models. Compared to the few-shot transfer performance shown in the \Tref{tab:main_results}, we can further observe: 1) the original supervised pretrained model has better few-shot transfer performance than the original unsupervised pretrained model. This is because supervised pretrained models are often trained with the cross entropy loss by using supervised labels, and there is not a similar distribution term in the cross entropy loss to encourage uniformity. Therefore, supervised pretrained representation is more compact; 2) by equipping with our unsupervised finetuning, the performance gap between unsupervised and supervised pretrained model becomes much smaller. Note that, we also try the naive unsupervised finetuning strategy upon the supervised pretrained model, but find its transfer performance is very bad, so we do not provide its results in the table.

\paragraph{Generalization Ability to Large Models.} Besides the pretraining method, our unsupevised finetuning strategy also generalizes well to large models. In the \Tref{tab:simple_overall_results_res101}, we take the supervised pretrained ResNet101 model as an example and evaluate its transfer performance. It shows that, even though the large model itself can produce better transfer performance than the default ResNet50 model, it can still improve its performance in a similar way to the small model by using the unsupervised finenetuned model as initialization.

\begin{table*}[t]
\small
  \begin{center}
  \setlength{\tabcolsep}{4.0mm}{
    \begin{tabular}{l|l|lllll}  
    \toprule  
   $K$ & Method 
   & Caltech101 & Pet   & CIFAR100  & Food101  & Avg \cr 
    \hline 
    \multirow{2}{*}{Semi-2 Acc} & Unsup-$S$ 
     & 79.04  & 54.52  & 41.43  & 18.23 & 48.31\cr
    &  + Our Unsupft-$T$ 
     & \reshl{82.84}{+}{3.80} & \reshl{68.58}{+}{14.06}  & \reshl{56.35}{+}{14.92}  &  \reshl{70.33}{+}{52.10} & \reshl{69.53}{+}{21.22}  \cr
    \hline
    \multirow{2}{*}{Semi-4 Acc} & Unsup-$S$  
     & 86.07  & 68.18   & 61.54  & 34.67 & 62.61 \cr
    &  + Our Unsupft-$T$ 
    & \reshl{86.87}{+}{0.8}    & \reshl{73.25}{+}{5.07}  & \reshl{66.18}{+}{4.64}   & \reshl{79.35}{+}{44.68}   & \reshl{76.41}{+}{13.80}\cr
    \hline
    \multirow{2}{*}{Semi-8 Acc} & Unsup-$S$  
       & 90.12 & 75.39   & 70.59 & 59.06 & 73.79 \cr
    &  + Our Unsupft-$T$ 
    & \reshl{90.27}{+}{0.15}  & \reshl{79.18}{+}{3.79}  & \reshl{71.65}{+}{1.06}    & \reshl{82.35}{+}{23.29} &  \reshl{80.86}{+}{7.07} \cr
\bottomrule  
    \end{tabular} }
    \captionof{table}{\small Semi-supervised learning results by using different models as initialization. ``Semi-$K$'' means $K$ labeled samples together with the whole unlabeled data are used for semi-supervised learning. Other annotations are similar as \Tref{tab:main_results}. It shows that, even though semi-supervised learning also leverage the unlabeled data, our unsupervised finetuned model can provide better initialization for it and boost the final performance. } 
    \label{tab:seemi_overall_results}
     \end{center}
    \vspace{-1em}
\end{table*}

\paragraph{Application in Semi-supervised Learning.} The essence of unsupervised finetuning is utilizing the unlabeled target data to get a better representation. Similarly, in  semi-supervised learning, it also aims to leverage the unlabeled target data as the extra learning guidance to achieve better performance. In this experiment, we are curious about whether semi-supervised learning can still benefit from unsupervised finetuned representation under the same data setting. To answer this question, we leverage the state-of-the-art semi-supervised learning method FixMatch \cite{sohn2020fixmatch}, and use the original unsupervised pretrained representation and unsupervised finetuned representation as the initialization respectively. 

Surprisingly, as shown in the \Tref{tab:seemi_overall_results}, semi-supervised learning also benefits from better initialization models from unsupervised finetuning, but the overall performance gain is smaller. In our understanding, many semi-supervised methods including FixMatch use the pseudo labels of the unlabeled data as the extra learning constraints, so providing better initialization will help generate more precise pseudo labels, subsequently improving the final performance. Conversely, if the initialization representation is not good, the pseudo labels will contain more errors and  the gain from the unlabeled data will be very marginal. For example, compared to the few-shot transfer, when only 2 labeled samples exist for each category on the Food101 dataset, semi-supervised learning initialized from the ImageNet unsupervised pretrained model can only bring about 2\% performance gain. By contrast, by using our unsupervised finetuned model as initialization, the extra performance gain from semi-supervised learning is 8.3\%.

\begin{table}[t]
\small
  \begin{center}
  \setlength{\tabcolsep}{1.0mm}{
    \begin{tabular}{l|ccc|ccc}  
    %\toprule  
   {Dataset}& Unsupft & SDR & Mix & $K=2$ & $K=4$ & $K=8$ \cr  
    \shline
    \hline
     \multirow{4}{*}{Caltech101}
     &--   &  &  & 63.12             & 76.54             & 84.45  \cr
     & \cmark &  &  & 68.48             & 78.18             & 82.03 \cr
     & \cmark & \cmark & & 75.28 & 82.99 & 85.15 \cr
     & \cmark & \cmark & \cmark & 81.02             & 85.35            & 87.24  \cr
     \hline
      \multirow{4}{*}{Pet}
     &--        &  &          & 56.84          & 68.94         & 75.10 \cr
     &  \cmark &  &         & 55.94          & 67.03        & 72.92 \cr
     & \cmark & \cmark &    & 62.81           & 71.35        & 75.14\cr
     &  \cmark & \cmark & \cmark & 70.29         & 76.89         & 79.88 \cr

     \hline
     \multirow{4}{*}{CIFAR100}
     & --  &  &  &    22.95          & 35.08             & 47.84  \cr
     & \cmark &  &  & 30.15             & 39.70             & 48.59 \cr
     & \cmark & \cmark & & 35.60 & 45.01 & 52.62 \cr
     & \cmark & \cmark & \cmark & 44.51             & 52.60            & 59.68  \cr
     
       \hline
     \multirow{4}{*}{Food101}
      & -- & &               & 15.84           & 25.49& 35.50\cr
      & \cmark  &  &       & 39.64           & 50.82& 59.58  \cr
      & \cmark & \cmark&   &  38.87               &   48.97             & 56.85\cr
      & \cmark  & \cmark & \cmark & 61.99        & 71.14       & 76.86  \cr
  
    \shline
    \end{tabular} }
    \captionof{table}{\small Ablation study of the two key strategies: sparse source data replaying (SDR) and data mixing (Mix). ``--'' means baseline without unsupervised finetuning,  ``\cmark'' means having the corresponding strategy. The second row of each dataset only with ``Unsupft'' checked denotes the naived unsupervised finetuning. It shows that both source data replaying and data mixing can help learn better representations. }
    \label{tab:source_replay}
     \end{center}
    \vspace{-2.5em}
\end{table}

\paragraph{Importance of Source Data Replaying and Data Mixing.} Sparse source data replaying and data mixing are the two keys in our unsupervised finetuning strategy. To quantitatively demonstrate their importance, we conduct the ablation studies by gradually removing each of them. As shown in the \Tref{tab:source_replay}, on all the datasets except Food101, source data replaying can boost the transfer performance by a large margin. For example, on the Caltech101 and Pet dataset, using source data replaying will boost the 2-shot transfer performance by 6.8\%, 6.9\% respectively. The reason why source data replaying cannot help Food101 should be because Food101 already has a larger number of images for each category, \ie, 750 images per category. As for data mixing, it consistently brings better performance on all the datasets. Especially on the Food101 dataset, the 2-shot transfer performance is significantly boosted by 23.12\%.

% \noindent \textbf{Transfer performance vs number of labeled samples}
\section{Conclusion}
Unsupervised learning has achieved great progress in recent years and become one of the most active and hot research topic. However, almost all existing works focus on the large-scale unsupervised pretraining, \ie, conduct unsupervised learning on a large-scale dataset. But in many real application scenarios, collecting such a large-scale unlabeled dataset is difficult or impossible. In this paper, we study ``unsupervised finetuning'', which aims to relax this requirement by unsupervised learning on a small-scale unlabel dataset on top of the pretrained representation. To the best of our knowledge, there is no prior work that focuses on this problem in the computer vision field yet. According to our analysis, we find unsupervised learning on the small-scale unlabeled data either from scratch or from the pretrained representation does not work well. After analyzing the widely used contrastive loss, we propose two simple and effective strategies, ``sparse source data replaying'' and ``data mixing''. Through extensive experiments, we demonstrate the proposed unsupervised finetuning strategy can achieve better performance than the naive unsupervised finetuning strategy. We hope our study will inspire more research in this direction.

{\small
\bibliographystyle{ieee_fullname}
\bibliography{egbib}
}

\end{document}